\pdfoutput=1

\documentclass[11pt]{article}

\usepackage[final]{acl}

\usepackage{times}
\usepackage{latexsym}

\usepackage[T1]{fontenc}

\usepackage[utf8]{inputenc}
\usepackage{multirow}
\usepackage{subfigure}

\usepackage{microtype}
\usepackage{amsfonts}       
\usepackage{amsmath}
\usepackage{amssymb}
\usepackage{mathtools}
\usepackage{amsthm}
\usepackage{caption}
\usepackage{enumitem}
\usepackage{caption}
\usepackage{wrapfig}
\usepackage{graphicx}
\usepackage{subfigure}
\usepackage{physics}

\usepackage{inconsolata}

\theoremstyle{plain}

\theoremstyle{definition}

\theoremstyle{remark}

\let\svthefootnote\thefootnote \newcommand\freefootnote[1]{%
  \let\thefootnote\relax%
  \footnotetext{#1}%
  \let\thefootnote\svthefootnote%
}

\usepackage{url}

\usepackage{pifont}

\title{MediSwift: Efficient Sparse Pre-trained Biomedical Language Models}

\author{Vithursan Thangarasa$^{\dagger}$ \quad Mahmoud Salem \quad
\textbf{Shreyas Saxena}$^{\text{*}}$ \\ \quad \textbf{Kevin Leong} \quad
\textbf{Joel Hestness} \quad \textbf{Sean Lie}\\
 Cerebras Systems \\ 
 \small{\texttt{\{vithu, sean\}}@cerebras.net} \\}

\begin{document}
\maketitle

\begin{abstract}
Large language models (LLMs) are typically trained on general source data for
various domains, but a recent surge in domain-specific LLMs has shown their
potential to outperform general-purpose models in domain-specific tasks (e.g.,
biomedicine). Although domain-specific pre-training enhances efficiency and
leads to smaller models, the computational costs of training these LLMs remain
high, posing budgeting challenges. We introduce MediSwift, a suite of biomedical
LMs that leverage sparse pre-training on domain-specific biomedical text data.
By inducing up to 75\% weight sparsity during the pre-training phase, MediSwift
achieves a 2-2.5x reduction in training FLOPs. Notably, all sparse pre-training
was performed on the Cerebras CS-2 system, which is specifically designed to
realize the acceleration benefits from unstructured weight sparsity, thereby
significantly enhancing the efficiency of the MediSwift models. Through
subsequent dense fine-tuning and strategic soft prompting, MediSwift models
outperform existing LLMs up to 7B parameters on biomedical tasks, setting new
benchmarks~w.r.t efficiency-accuracy on tasks such as PubMedQA. Our results show
that sparse pre-training, along with dense fine-tuning and soft prompting,
offers an effective method for creating high-performing, computationally
efficient models in specialized domains.\freefootnote{$^{\text{*}}\text{Work
done while at Cerebras.}$} \freefootnote{$^{\dagger}\text{Corresponding
author.}$}
\end{abstract}

\vspace{-10pt}
\section{Introduction}

\begin{figure}[ht]
  \begin{center}
  \centerline{\includegraphics[width=1.0\columnwidth]{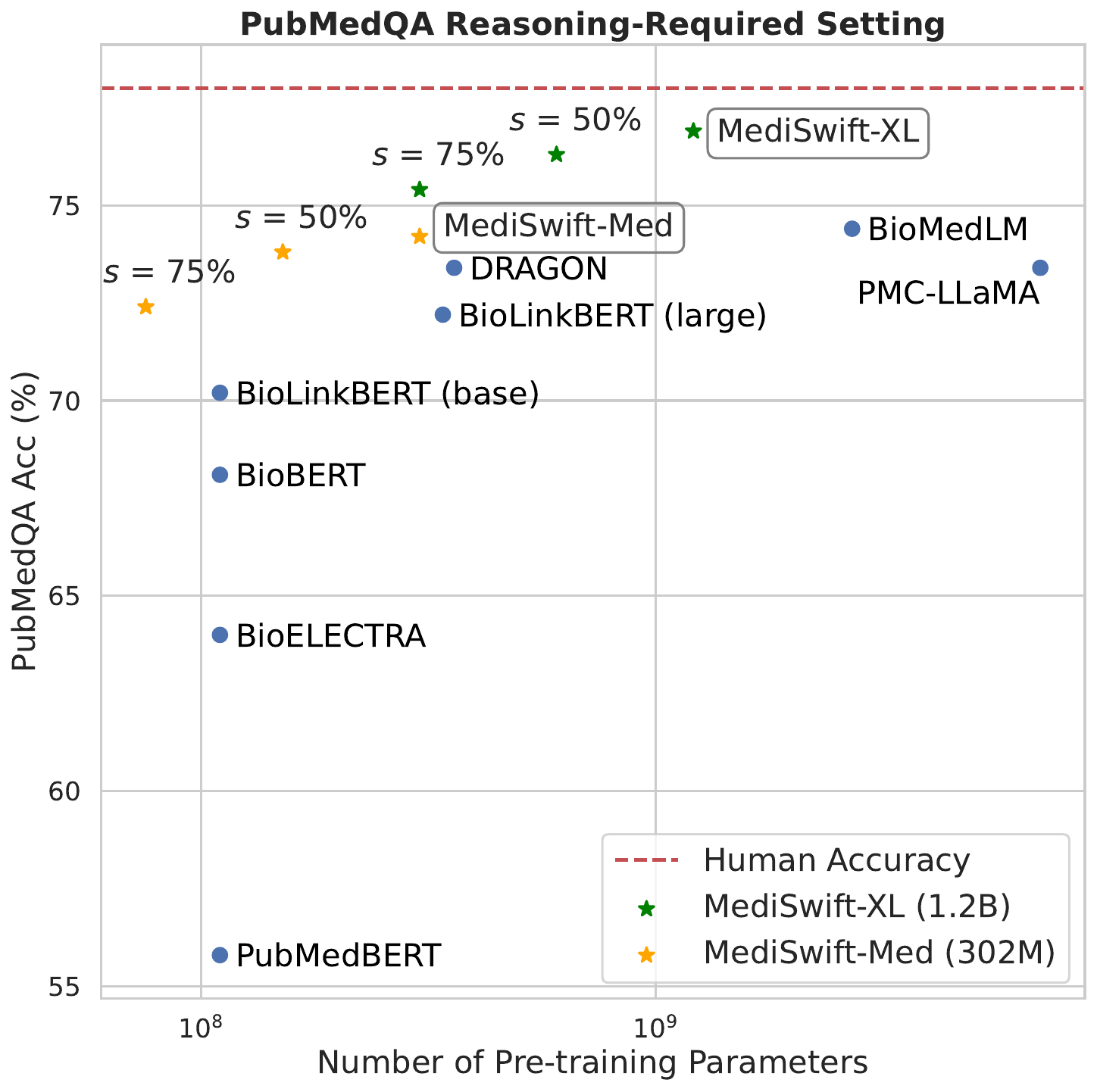}}
  \caption{\textbf{Comparison of Model Size vs. PubMedQA Accuracy in the
  Reasoning-Required Setting:} Our dense and sparse MediSwift models noticeably
  outperform other fine-tuned language models $\leq$ 7B parameters, improving
  the efficiency-accuracy pareto frontier. In particular, MediSwift-XL (1.21B)
  achieves new state-of-the-art 76.8\% accuracy at this size (i.e., being 5.8x
  smaller than PMC-LlaMA). In addition, sparse pre-trained MediSwift-XL models
  at $s \in \{50\%, 75\%\}$ outperform other models at similar or larger size.
  Additional details are provided in Table~\ref{tab:mediswift_pretrain_losses}
  and~\ref{tab:mediswift_pubmed}.}
  \label{fig:mediswift_pubmedqa}
  \vspace{-15pt}
  \end{center}
\end{figure}

The landscape of large language models (LLMs) has been predominantly shaped by
efforts aimed at creating generalized models~\citep{openai2023gpt4,
touvron2023llama, zhang2022opt, shoeybi2020megatronlm}, trained on diverse
datasets that encompass a wide array of topics and domains, such as
Pile~\citep{gao2020pile}, MassiveText~\citep{hoffmann2022an} and
RedPajama~\citep{together2023redpajama}. While these comprehensive datasets have
included data in specialized domains (e.g., programming
code~\citep{chen2021evaluating} and PubMed Central~\citep{gao2020pile}), the
overarching goal has been to forge LLMs capable of broad applicability. However,
recent efforts in training models on domain-specific data are emerging, with
these smaller, specialized models surpassing general-purpose ones in domain
focused tasks, especially in science~\citep{taylor2022galactica} and
medicine~\citep{luobiogpt, elliotbiomedlm}. This has sparked a renewed interest
in the development of LLMs tailored to specific domains, suggesting a promising
avenue for enhancing compute efficiency and model performance~w.r.t evaluation
metrics on downstream tasks.

Furthermore, the shift towards specialized LLMs in the field of medicine is
particularly gaining recognition for its capacity to significantly improve the
accuracy and effectiveness of these models. This is achieved by closely aligning
them with the specific needs and complexities of this specialized area. For
example, in biomedicine, models trained on specialized literature (e.g.,
MEDITRON~\citep{chen2023meditron70b}, BioGPT~\citep{luobiogpt},
BioMedLM~\citep{elliotbiomedlm}) exhibit significant improvements over general
ones. This approach involves either continued pre-training on domain-relevant
texts or pre-training models from scratch with domain-specific data (e.g.,
PubMed Central\footnote{https://pubmed.ncbi.nlm.nih.gov/\label{pubmed}} and
PubMed open-access research papers), emphasizing the accuracy benefits and
ability to yield models that are more compute efficient. 

After pre-training models on domain-specific data, further accuracy enhancements
are achieved through prompt-based fine-tuning~\citep{nori2023generalist,
reynolds2021prompt, Peng2023ASO, liu-etal-2022-p} and full fine-tuning of the
weights~\citep{ziegler2019finetuning, lamda2022, hu2021lora}, offering a balance
between adaptability, efficiency, and task-specific optimization. This dual
approach merges the model's pre-trained knowledge with the specific needs of
downstream tasks, thereby facilitating a more accurate and effective application
of the model's capabilities across a wide array of downstream applications;
maximizing the potential of domain-specific LLMs.

Although domain-specific LLM pre-training offers significant benefits, the
computational demands remain a significant challenge, necessitating innovative
solutions for broader accessibility and efficiency. Techniques proposed to
mitigate these burdens include sparse attention~\citep{dao2022flashattention,
jaszczur2021sparse}, quantized optimization~\citep{tang20211bit}, low-rank
factorization~\citep{lialin2023relora}, and sequence-level curriculum
learning~\citep{li2022the}. Sparse attention enhances transformer efficiency but
mainly benefits attention layers and may struggle with long-range dependencies.
Quantized optimization reduces memory footprint and accelerates training with
lower-precision arithmetic but introduces quantization noise and relies on
hardware support. Low-rank factorization saves memory and computation by
approximating weight matrices, risking information loss and requiring
architectural changes. Sequence-level curriculum learning improves convergence
and generalization by progressively increasing difficulty, primarily enhancing
training efficiency. Among these approaches, weight sparsity emerges as a
promising method, distinct from the others in its approach by setting a subset
of model parameters to zero, thus reducing the FLOPs needed during training.
Unlike sparse attention, which optimizes a specific component, or quantized
optimization and low-rank factorization, which may introduce trade-offs in model
performance, sparse pre-training directly reduces computational costs across the
entire model without inherently compromising performance.

However, the adoption of sparse training is limited by (1) the challenge of
finding optimal sparsity patterns which retain the accuracy of the original
dense model~\citep{frankle2018lottery, ma2022effective} and (2) its difficulties
in accelerating on hardware optimized for dense computations (e.g., Nivida GPUs
and Google TPUs)~\citep{2020shooker}. Additionally, sparse pre-training of LLMs
typically leads to an accuracy loss from optimization challenges in sparse
networks~\citep{Evci2019TheDO}, but previous studies have shown that
transitioning from sparse-to-dense training can effectively address these
problems~\citep{thangarasa2023spdf,dao2022monarch}. Our work aims to reduce
training FLOPs by employing unstructured weight sparsity in domain-specific LLMs
and recovering lost representational capacity by transitioning to dense weight
matrices for downstream fine-tuning.

Despite weight sparsity not being widely used in real-world applications,
advancements in specialized software
kernels~\citep{gale2019state,elsen2019sparse,Ashby2019ExploitingUS,
tangandyang2023torchsparse} have facilitated inference acceleration with
unstructured sparsity. Recent developments have shown that the benefits of
unstructured weight sparsity can be fully harnessed on specialized hardware,
such as the Cerebras CS-2~\citep{lie_2023, cerebrasHarnessingPower}, for LLM
training. As sparse training techniques and hardware continue to co-evolve, we
anticipate that the reductions in FLOPs will lead to realized sparse
acceleration. The latest innovations in software and
hardware~\citep{neural_magic_2021, abhaysparse2024} are geared towards enabling
the widespread adoption of unstructured weight sparsity, offering the potential
to achieve higher compression ratios and practical speedups in terms of
wall-clock time. 

Building on this momentum, we introduce MediSwift, a suite of biomedical
language models (LMs) available in three sizes: Med (302M), Large (510M), and XL
(1.21B). These models are based on GPT-3 and pre-trained sparsely from scratch on
biomedical texts, aimed at reducing the computational costs required for
training. We explore the impact of applying 50\% and 75\% weight sparsity during
pre-training, which results in a 2-2.5x reduction in the overall training FLOPs
needed. Figure~\ref{fig:mediswift_pubmedqa} summarizes the performance, where
our dense and sparse MediSwift models noticeably outperform other language
models up to 7B parameters. We demonstrate MediSwift's capabilities through
fine-tuning on established benchmarks for biomedical natural language processing
(NLP) tasks (e.g., PubMedQA~\citep{jin2019pubmedqa} for question answering and
HoC~\citep{Baker2016AutomaticSC} for document classification) showing
significant improvements in the balance between efficiency and accuracy.

Although previous research suggests that sparse pre-training may compromise
model accuracy on downstream tasks, our approach incorporates dense fine-tuning
with strategic soft prompting to effectively regain performance on specialized
tasks. Specifically, MediSwift-XL (1.21B) sets a new state-of-the-art by reaching
76.8\% accuracy, despite being 5.8x smaller than PMC-LlaMA. Moreover,
MediSwift-XL models, pre-trained with 50\% and 75\% sparsity, surpass the
performance of models of similar or greater sizes (e.g., MediSwift-XL at 75\%
sparsity outperforms BioMedLM while being almost 9x smaller). Our work not only
highlights the potential for sparse pre-training to make LM training more
economically viable but also sets a new benchmark for efficiency in
domain-specific applications of LLMs. The key contributions are:

\begin{enumerate}
  \item We introduce MediSwift, a family of biomedical language models in three
  sizes (Med, Large, and XL), and extend this by introducing both dense and
  sparse variants pre-trained with 50\% and 75\% weight sparsity. This
  diversification balances computational efficiency with model effectiveness in
  biomedical applications, offering options for different computational resource
  needs.
  \item To our knowledge, this is the first study to highlight the benefits of
  sparse pre-training on biomedical texts, achieving significant computational
  savings. We show inducing 75\% weight sparsity into MediSwift models, results
  in a 2.5x reduction in training FLOPs, while improving efficiency-accuracy
  trade-offs in tasks like PubMedQA. 
  \item We demonstrate that despite the potential for sparse pre-training to
  reduce model accuracy, dense fine-tuning combined with soft prompting can
  effectively regain performance on task-specific fine-tuning. Specifically,
  50\% sparse MediSwift-XL achieves a new state-of-the-art with 76.3\% accuracy
  on PubMedQA, surpassing existing models up to 7B parameters.
\end{enumerate}

\vspace{-5pt}
\section{Methodology}
In this section, we formalize our two-phase training framework for MediSwift
models to reduce computational costs and yet retain model accuracy. Initially,
we pre-train these models on biomedical data, applying unstructured weight
sparsity to reduce the computational training FLOPs. Following this, we enhance
the model through dense fine-tuning, reactivating weights to improve
adaptability for specific tasks, and incorporate soft prompting to refine
responses for task requirements. This efficient approach, combining sparse
pre-training, dense fine-tuning, and soft prompting, significantly boosts both
model efficiency and performance, as our results demonstrate in
Section~\ref{sec:results}.

\subsection{Autoregressive Language Modeling}
\paragraph{Dense Pre-training} Autoregressive LMs predict a series of tokens by
making each token's prediction dependent on the ones before it, similar to a
Markov chain process. This method follows core principles of language modeling,
aiming to understand the pattern of token sequences unsupervisedly from a corpus
of text data. Consider an unsupervised corpus $\mathcal{U}$ consisting of tokens
${u_1, u_2, \ldots, u_{|\mathcal{U}|}}$, with $|\mathcal{U}|$ denoting the
corpus's total token count. Our objective is to enhance the model's ability to
predict sequences by maximizing the likelihood of the observed sequences,
formulated as the sum of the log probabilities of each token given its preceding
context within a window of size $k$. The mathematical representation of this
objective is as follows:
\begin{equation}
  \label{eq:gpt}
  \mathcal{L}(\mathcal{U}) = \sum_{i=1}^{|\mathcal{U}|}\log(p(u_i | u_{i-k:i-1}, \theta)),\notag
\end{equation}
where $\theta$ denotes the neural network's parameters, encapsulating the
\textit{dense} configuration of the network's architecture. The context window,
$k$, determines the number of preceding tokens used for current token
prediction. The neural network, parameterized by $\theta \in \mathbb{R}^{N}$,
where $N$ is the total parameter count, aims to optimize these parameters across
all layers $L$, with each layer $l$ having its own set of parameters $\theta_l$.

\vspace{-5pt}
\paragraph{Sparse Pre-training} Building upon our framework for dense
pre-training, we introduce weight sparsity into the model, specifically to
improve the computational efficiency. We achieve this by methodically reducing
the number of active connections within each layer $l$ of the model by a
predefined sparsity level $s_l \in (0, 1)$, effectively rendering a portion of
the network's parameters inactive. This process yields a network with
$(1-s_l)N_l$ active parameters per layer, where $N_l$ denotes the original
number of parameters in layer $l$. The overall sparsity is quantified by $S$,
which represents the ratio of inactive parameters to the total parameter count
of the initially dense model, calculated as $S = \frac{\sum_l^L s_lN_l}{N}$. To
apply sparsity effectively, we employ a binary mask $m \in \{0,1\}^{|\theta|}$
to the model's initial parameters $\theta^0$, resulting in a sparse parameter
set $m \odot \theta^{0}$. Our approach to inducing sparsity involves random
parameter pruning, a process where $S$ percentage of the model's weights are
randomly set to zero at initialization. This mask effectively segregates the
parameters into active (1) and inactive (0) states, thereby establishing a
sparsity-induced version of the language model that aims to minimize a slightly
modified objective:
\begin{equation}
  \label{eq:gpt}
  \mathcal{L}(\mathcal{U}) = \sum_{i=1}^{|\mathcal{U}|}\log(p(u_i | u_{i-k:i-1}, m \odot \theta)).
\end{equation}
We leverage the GPT-3~\citep{brown2020gpt3} architecture for the MediSwift
biomedical language model, training it with the
AdamW~\citep{loshchilov2017decoupled} optimizer on a curated biomedical dataset,
following the objective shown in Eq.~\ref{eq:gpt} for $j$ iterations to obtain
parameters $\theta^j$. This pre-trained model is then fine-tuned for specific
supervised tasks in the biomedical domain. GPT-3 was selected for MediSwift due
to its versatility in NLP tasks and suitability for specialized domains like
biomedicine. Its autoregressive nature and capability to produce contextually
relevant text align with our methodology of sparse pre-training, dense
fine-tuning, and soft prompting to enhance efficiency and task performance.
Additionally, GPT-3 was well-supported on the Cerebras CS-2 hardware, with
optimized kernels for sparse training using unstructured weight sparsity,
further enhancing our model's computational efficiency.

Our approach is fundamentally model-agnostic and adaptable to various LLMs. The
core components of our methodology (i.e., sparse pre-training to reduce
computational burden, followed by dense fine-tuning and soft prompting to regain
or enhance task performance) can be broadly applied across different LLMs,
including newer iterations of GPT.

\subsection{Dense Fine-tuning and Soft Prompting}
In this section, we detail the adaptation of our pre-trained MediSwift model for
tasks like biomedical question answering (QA) and document classification using
dense fine-tuning and soft prompting. We align tasks with varying output formats
to our pre-training format by converting task labels into natural language
sequences~\citep{li-liang-2021-prefix, hu2021lora}. This method avoids
structured formats and special tokens, ensuring semantic coherence and making
full use of the natural language corpus MediSwift was trained on.

Following~\citet{luobiogpt}, each downstream fine-tuning task is represented by
a training set consisting of \texttt{source-target} pairs defined as:
$\mathcal{Z} = \{(x_1 , y_1),(x_2 , y_2),\ldots,(x_{|x|}, y_{|y|})\}$, where
both $x$ and $y$ are sequences of tokens. For example, in question answering
(e.g., PubMedQA), $x$ corresponds to the question and reference context
description, and $y$ the categorical answer to the question; in biomedical
document classification (e.g.,  Hallmarks of Cancers corpus), $x$ is the text
passage and $y$ corresponds to the hallmarks of cancer.

\subsubsection{Dense Fine-tuning}\label{subsec:denseft} We begin fine-tuning
with parameters $\theta^j$ set at their pre-trained values, adjusting them by a
task-specific increment $\Delta\theta$ with the same dimensionality,
$|\Delta\theta| = |\theta|$. Unlike prior efforts that aimed at parameter
efficiency for easier model deployment~\citep{hu2021lora, dettmers2023qlora}, we
prioritize reducing pre-training computational costs via unstructured weight
sparsity and enhance network representation by adopting dense
fine-tuning~\citep{thangarasa2023spdf}. This approach overcomes sparse
optimization challenges by reactivating previously inactive weights during the
dense fine-tuning phase, thus enhancing the model's capacity. By removing the
sparsity mask $m$, we allow $\sum_{l}^{L}s_l {\cdot} N_l$  weights to be
reactivated and initialize them to zero—a method proven more effective than
other initialization strategies (e.g., scaled uniform or normal distribution
initializations~\citep{evci2020rigging}). The network is then densely updated to
optimize the loss function:
\begin{align}
  \mathcal{L}(\mathcal{Z}) &= \sum_{(x,y)\in\mathcal{Z}} \log \prod_{t=1}^{|\boldsymbol{y}|} \label{eq:densefinetune} \\
  &\quad p(y_t | (x_{1},\ldots,x_{t-1}), \theta^j + \Delta\theta) \notag
\end{align}

\subsubsection{Soft Prompting}\label{subsec:prompting} Previous research on
language models in the biomedical field has mainly focused on fine-tuning for
domain-specific tasks~\citep{elliotbiomedlm}. Recently, there has been a move
towards improving biomedical NLP task performance through prompt
engineering~\citep{luobiogpt,nori2023generalist, yagnik2024medlm}. Drawing
inspiration from these advancements, our approach integrates prompt-based
techniques into the fine-tuning phase of MediSwift. More precisely, we adopt the
soft prompting methodology as described by~\citet{luobiogpt}, aiming to refine
our model's capability in understanding and processing biomedical text. Similar
to existing work on prompt tuning~\citep{brown2020language, liu2021gpt,
lester-etal-2021-power}, to integrate soft prompts, we insert virtual tokens
between the source and target sequences, thus modifying the loss function
$\mathcal{L}(\mathcal{Z})$ in Eq.~\ref{eq:densefinetune}. This adjustment
accounts for the \texttt{[source; prompt; target]} sequence structure, impacting
the model's learning and inference. Let $\mathcal{P}$ denote the prompt,
consisting of a sequence of virtual tokens $\in \{v_1, v_2,\ldots,v_n\}$, where
$n$ is the number of virtual tokens, and these tokens are represented by
continuous embeddings. The modified Eq.~\ref{eq:densefinetune}, reflecting the
inclusion of the prompt and its positioning, can be formalized as:
\begin{align}
  \mathcal{L}(\mathcal{Z}) &= \sum_{(x,y)\in\mathcal{Z}} \log \prod_{t=1}^{|\boldsymbol{y}|} \label{eq:softprompt} \\
  &\quad p(y_t | ([x; \mathcal{P}]; y_{<t}), \theta^j + \Delta\theta) \notag
\end{align}

Through this multi-faceted approach, our pre-training and fine-tuneing method
for MediSwift not only addresses the computational efficiency challenges in LM
training, but also leverages the capabilities of in-domain pre-trained LMs to
improve performance on biomedical NLP tasks.

\section{MediSwift Biomedical Pre-training}

This section describes the MediSwift pre-training process, including data
sources, collection, and preprocessing for biomedical data. We explain the
dataset's origins, statistical analysis, and preparation for efficient training.
We also compare MediSwift models' performance with both dense and sparse
pre-training, emphasizing training convergence differences and FLOPs savings. 

\vspace{-5pt}
\subsection{PubMed Papers and Abstracts}\label{sec:pretraincorpa} MediSwift is
an in-domain biomedical language model, drawing its strength from an exclusive
pre-training regimen focused solely on biomedical textual data. Its foundation
lies in the vast repository of available open-access medical research papers and
abstracts found in PubMed Central (PMC)~\citep{pubmedcentral}, similar to the
approaches used in prior models (e.g., Meditron~\citep{chen2023meditron70b},
BioGPT~\citep{luobiogpt}, BioMedLM~\citep{elliotbiomedlm}). 

PMC consists of 4.91M full-text papers, and PubMed and PMC Abstracts comprise of
16.1 million papers (see Table~\ref{tab:datasets}). Moreover, we gathered the
most recent PubMed entries, updated prior to 2023, directly from the official
website$^{\ref{pubmed}}$, utilizing the official scripts for PubMed
Abstracts\footnote{\url{https://github.com/thoppe/The-Pile-PubMed?tab=readme-ov-file}}
and PubMed
Central\footnote{\url{https://github.com/EleutherAI/pile-pubmedcentral}}.
Similar to~\citet{luobiogpt} and~\citet{chen2023meditron70b}, we filter out
empty items containing solely titles without accompanying abstracts.
Furthermore, prior works have show the significance of in-domain vocabulary for
improving performance of specialized
LMs~\citep{yupubmedbert,wu2023bloomberggpt,mielke2021words}, a critical step
that is often overlooked. Therefore, inspired by~\citet{luobiogpt}
and~\citep{elliotbiomedlm}, instead of using the standard GPT-3 vocabulary, we
learned the vocabulary directly from the biomedical corpus. Employing
Moses~\citep{koehn-etal-2007-moses} tokenization followed by byte pair encoding
(BPE), we segment the corpus into word pieces and learn the vocabulary;
resulting in a size of 42,384. By exclusively pre-training with biomedical texts
and using a specialized vocabulary, MediSwift improves the efficiency-accuracy
frontier, as empirically shown on biomedical tasks in Section~\ref{sec:results}.

\begin{table}[!t]
  \begin{center}
  \caption{Statistics on the mixture of pre-training data for MediSwift,
  including the sizes of the training and validation sets. Total sample count is
  provided for each set, along with the percentage of validation set allocation
  relative to the training set.}
  \begin{tabular}{ccc}
  \hline
  \multicolumn{1}{c}{\multirow{2}{*}{\textbf{Dataset}}} &
  \multicolumn{2}{c}{\textbf{Number of Samples}} \\ \cline{2-3}
  \multicolumn{1}{c}{}                                  & Train               &
  Validation               \\ \hline
  PubMed Abstracts$^{\ref{pubmed}}$                                      & 15.7M
  & 487K (3\%)               \\
  PubMed Papers$^{\ref{pubmed}}$                                          & 4.9M
  & 142K (3\%)               \\ \hline \hline
  \textbf{Total}                                                 & 20.6M & 629K
  \\ \hline           
  \end{tabular}
  \label{tab:datasets}
  \end{center}
  \vspace{-5pt}
\end{table}

\subsection{MediSwift Pre-training }

\paragraph{Pre-training Experimental Details} We pre-train and benchmarked
MediSwift in-domain biomedical language models at 3 sizes: 302M, 510M and 1.21B.
All MediSwift models are pre-trained from scratch using the Cerebras
CS-2\footnote{\url{https://docs.cerebras.net/en/2.1.1/wsc/how_to_guides/sparsity.html}},
taking advantage of its ability to accelerate training with unstructured
sparsity. At the time of the study, the specialized kernels of Cerebras CS-2
were designed to facilitate training with static unstructured sparsity (refer to
Appendix~\ref{appsec:cs2} for additional details). In the pre-training phase of
the MediSwift models,  50\%, and 75\% sparsity levels are explored, aside from
their respective dense counterparts. The pre-training of MediSwift models are
conducted on a single CS-2 for a total of 200,000 steps, with a batch size of
512 and a maximum sequence length of 1024 tokens, resulting in approximately
104.86B tokens processed in total (see Appendix~\ref{appsub:pretrain} for
additional pre-training experimental setup details). 

\subsection{Sparse Pre-trained MediSwift}
While there are several advanced sparse training
techniques~\citep{evci2020rigging,mocanu2018, liu2021sparse}, for simplicity, in
this work, we adopt static sparsity, namely random pruning, for the sparse
pre-training of MediSwift models. This approach mandates a uniform distribution
of sparsity levels throughout each layer, irrespective of the layer's parameter
count or its FLOPs. Specifically, the scope of our sparsification process is
confined to all dense linear layers within the network, including both matrices
within the multi-layer perceptron (MLP) module—namely, the intermediate layer
and the MLP output projection, as well as the four weight matrices integral to
the self-attention~\citep{vaswani2017transformers} mechanism: query, key, value,
and attention output projection. Notably, we ensure that the embeddings, Layer
Normalization~\citep{ba2016layer} components and biases are kept dense. 

\begin{figure}[t]
      \centering
      \includegraphics[width=\linewidth]{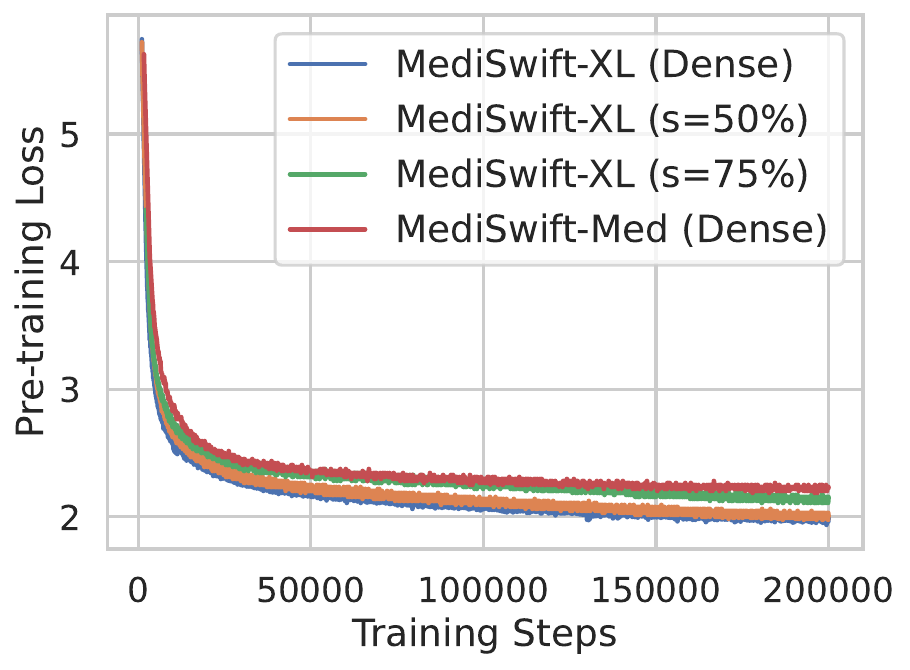}
      \caption{\textbf{Comparison of pre-training loss curves for MediSwift
      models}: MediSwift-XL's training loss reveals that at 50\% sparsity, the
      model's performance closely mirrors that of its dense variant, with
      negligible effects on training loss. At 75\% sparsity, although the gap in
      training loss widens, the sparse MediSwift-XL still outperforms the dense
      MediSwift-Med, showcasing efficient learning even at higher sparsity
      levels.}
      \label{fig:mediswiftxl_pretrain_losses}
  \label{fig:main}
\end{figure}

\begin{table}[t]
  \centering
  \caption{\textbf{Final Pre-training Losses and Computational Efficiency of
  MediSwift Models.} We summarize the results for the MediSwift-Med and
  MediSwift-XL models, trained using the biomedical pre-training corpus detailed
  in Section~\ref{sec:pretraincorpa}. We report the final pre-training losses
  for dense and sparse variants at 50\% and 75\% sparsity levels. The total
  FLOPs and FLOP savings relative to the dense baseline are indicated in
  parentheses, highlighting the models' computational efficiency.}
  \label{tab:mediswift_pretrain_losses}
  \begin{tabular}{lccc}
    \hline
    \textbf{MediSwift} & \textbf{Size} &
    \textbf{\begin{tabular}[c]{@{}c@{}}Pre-train \\ Loss\end{tabular}} &
    \textbf{\begin{tabular}[c]{@{}c@{}}Train \\ FLOPs (x 10$^{20}$)
    \end{tabular}} \\ \hline
    Med$_{\text{dense}}$ & 302M & 2.234 & 2.677 (1.00x) \\
    Med$_{s=50\%}$ & 151M & 2.265 & 1.727 (0.64x) \\
    Med$_{s=75\%}$ & 0.76M & 2.375 & 1.252 (0.46x) \\ \hline
    \hline
    XL$_{\text{dense}}$ & 1.21B & 1.979 & 9.148 (1.00x) \\
    XL$_{s=50\%}$ & 605M & 2.012 & 5.348 (0.58x) \\
    XL$_{s=75\%}$ & 302M & 2.141 & 3.448 (0.38x)                \\ \hline                                  
    \end{tabular}
    \vspace{-5pt}
\end{table}

\vspace{-5pt}
\subsection{Analysis on Pre-trained Models} In
Figure~\ref{fig:mediswiftxl_pretrain_losses}, we illustrate the training loss
curves for the MediSwift-XL model for both the dense and sparse configurations
at $s \in \{50\%, 75\%\}$. The 50\% sparse model's training loss closely follows
that of the dense MediSwift-XL, showcasing minimal deviation throughout the
training process. However, a noticeable divergence is observed at 75\% sparsity,
where the final training loss slightly lags behind that of the dense
counterpart. Interestingly, when comparing the 75\% sparse MediSwift-XL model
with the dense MediSwift-Med model, the former, despite sharing the same number
of non-embedding parameters, it achieves a lower training loss (refer to
Table~\ref{tab:mediswift_pretrain_losses}). This observation aligns with
previous findings that larger, albeit sparser, models can surpass their smaller,
densely parameterized equivalents in terms of
performance~\citep{thangarasa2023sparse,liu2022unreasonable, ramanujan2020s,
golubeva2021are}. This superiority is further supported by the improved accuracy
of the 75\% sparse model on  biomedical NLP tasks, when compared to the
MediSwift-Med dense model (see Section~\ref{sec:results}), highlightling the
benefits of training larger sparse models in comparison to smaller dense ones.

In Table~\ref{tab:mediswift_pretrain_losses}, we provide an analysis of the
computational efficiency achieved through sparse pre-training in our proposed
MediSwift architectures, namely MediSwift-Med and MediSwift-XL. Here, we
quantify the total FLOPs required for both forward and backward propagations
during the pre-training phase of these models. For the MediSwift-XL model,
attention and vocabulary embeddings represent 13.3\% and 6.8\% of total FLOPs,
respectively, hence highlighting the computational savings at the 1.21B parameter
scale. Sparse pre-training with 75\% sparsity reduces FLOPs by slightly over
2.5x compared to its dense counterpart. The smaller MediSwift-Med model has a
higher FLOP percentage for attention and embeddings, hence achieving a 2x
reduction at the same sparsity level. This indicates that FLOP savings increase
with model size, demonstrating that larger models~\citep{hoffmann2022an,
kaplan2020scaling, hestness2017deep} can potentially benefit more from sparse
pre-training. In addition, we emphasize that the total FLOPs required for
fine-tuning these models account for a minor fraction of the overall
pre-training FLOPs, reinforcing the efficiency of our approach in scaling to
larger model sizes while conserving computational resources.

\vspace{-5pt}
\section{Fine-tuning on Biomedical tasks} \label{sec:results}

This section evaluates MediSwift's performance on PubMedQA and HoC benchmarks
using dense fine-tuning on each variant's specific training set (e.g.,
fine-tuning and testing with the PubMedQA dataset). During the fine-tuning
phase, the total training FLOPs constitute only a minor fraction of the FLOPs
expended during pre-training, despite the fine-tuning being conducted densely.
As a result, the FLOPs consumed during sparse pre-training is proportional to
the combined FLOPs of sparse pre-training and dense fine-tuning.
Following~\citet{luobiogpt}, we incorporate soft prompting into our fine-tuning
framework by formatting sequences as \texttt{[source; prompt; target]}. This
format helps our models better utilize contextual information, demonstrating
MediSwift's effectiveness and adaptability in medical text analysis. Further
details on hyperparameters and dataset specifics are provided in
Appendix~\ref{appsec:pubmedqa} and~\ref{appsec:hoc}. We note that all
fine-tuning results were averaged across 3 random seeds.

\begin{table}[!t]
  \caption{MediSwift's performance on the PubMedQA reasoning-required task in
  both dense and sparse settings, $s \in \{50\%, 75\%\}$. This table compares
  MediSwift against other language models from the PubMedQA leaderboard,
  demonstrating its efficiency-accuracy improvement. Results are shown for
  models $\leq$ 7B parameters, with the ``size'' column indicating pre-training
  parameters and the final column reporting test accuracy on the PQA-L test
  set.}
  \label{tab:mediswift_pubmed}
  \begin{tabular}{l@{\hspace{0pt}}cc}
    \hline
    \multicolumn{1}{c}{\textbf{Model}}                     &
    \multicolumn{1}{c}{\textbf{Size}}    & \multicolumn{1}{c}{\textbf{Acc.}}  \\
    \hline
  PubMedBERT$_{\text{\citep{yupubmedbert}}}$             &  110M & 55.8 \\
  BioELECTRa$_{\text{\citep{kanakarajan-etal-2021-bioelectra}}}$               &
  110M                                       & 64.2 \\
  BioLinkBERT$_{\text{base,~\citep{yasunaga-etal-2022-linkbert}}}$    & 110M &
  70.2                              \\
  BioLinkBERT$_{\text{large,~\citep{yasunaga-etal-2022-linkbert}}}$   & 340M &
  72.2                              \\
  BioGPT$_{\text{med,~\citep{luobiogpt}}}$ & 345M & 73.6$^{\dagger}$ \\
  DRAGON$_{\text{\citep{yasunaga2022dragon}}}$                   & 360M & 73.4
  \\ \hline
  MediSwift-Med$_{\text{dense}}$             & 302M & 74.2 \\
  MediSwift-Med$_{(s=50\%)}$   & 151M                                       &
  73.8                              \\
  MediSwift-Med$_{(s=75\%)}$   & 0.76M                                      &
  72.4                               \\ \hline \hline
  BioGPT$_{\text{large,~\citep{luobiogpt}}}$  & 1.54B & 75.5$^{\dagger}$ \\
  BioMedLM$_{\text{\citep{elliotbiomedlm}}}$                 & 2.70B & 74.4 \\
  PMC-Llama$_{\text{\citep{wu2023pmcllama}}}$                & 7.00B & 73.4 \\
  GPT-3.5 (0-shot)$_{\text{\citep{nori2023capabilities}}}$                    &
  - & 71.6 \\
  GPT-4 (0-shot)$_{\text{\citep{nori2023capabilities}}}$                    & -
  & 75.2 \\ \hline
  MediSwift-Large$_{\text{dense}}$           & 510M & 75.1 \\
  MediSwift-Large$_{(s=50\%)}$ & 255M                                       &
  74.2                              \\
  MediSwift-Large$_{(s=75\%)}$ & 128M                                       &
  73.4                             \\
  MediSwift-XL$_{\text{dense}}$              & \textbf{1.21B} & \textbf{76.8} \\ 
  MediSwift-XL$_{(s=50\%)}$    & 605M                                       &
  76.3                              \\
  MediSwift-XL$_{(s=75\%)}$    & 302M                                       &
  75.4                           \\ \hline  
  \end{tabular}
  \small $\dagger$ We followed the fine-tuning steps used in the official BioGPT
  code\footnotemark \ to reproduce the results on BioGPT$_{\text{med}}$ and
  BioGPT$_{\text{large}}$, which reported accuracies of 78.2\% and 81.0\%,
  respectively~\citep{luobiogpt}. However, the methodologies for fine-tuning on
  PubMedQA~\citep{luobiogpt}, as well as the fine-tuning scripts, lack clear
  descriptions and details, making it difficult to reproduce these results,
  especially under a \textit{reasoning-required setting}. Hence, we made efforts
  to replicate their findings as closely as possible, despite uncertainties
  about the original experimental setup.
  \vspace{-10pt}
\end{table}
\footnotetext{\url{https://github.com/microsoft/BioGPT/tree/main/examples/QA-PubMedQA}}

\vspace{-5pt}
\subsection{Question Answering with PubmedQA}\label{subsec:pubmedqa} We assess
MediSwift's performance on the PubMedQA~\citep{jin2019pubmedqa} dataset, which
is derived from PubMed abstracts and includes three subsets: PQA-A, PQA-U, and
PQA-L. We adhere to the original train/val/test splits, focusing on the PQA-L
test set for the final evaluation. Our approach utilizes multi-stage fine-tuning
and soft prompting with continuous embeddings of length $n$ = 9, which was shown
to perform the best in terms of accuracy~\citep{luobiogpt}. Following
established preprocessing methods, we format the data into \texttt{[source,
target]} sequences, each consisting of a question, reference context, long
answer, and a categorical label, \texttt{[yes/no/maybe]}, for the answer. The
performance is measured by classification accuracy, particularly under the
challenging \textit{reasoning-required setting}~\citep{jin2019pubmedqa}, where
the model predicts based on the question and context without the long answer.

In Table~\ref{tab:mediswift_pubmed}, we demonstrate that across all sizes,
MediSwift improves the pareto frontier in PubMedQA accuracy, notably with the
dense MediSwift-XL model setting a new benchmark while being significantly
smaller, at 5.8x less size than PMC-LLaMA. This trend continues with the 50\%
and 75\% sparse variants of MediSwift, which surpass other language models of
comparable or larger sizes. Specifically, the 75\% sparse MediSwift-XL exceeds
BioMedLM's performance by 1.0\% while being approximately 8.9x smaller.
Furthermore, within the MediSwift family, the larger yet sparse 75\%
MediSwift-XL demonstrates superior performance over the smaller dense
MediSwift-Med by 1.2\%, despite both models sharing the same pre-training
parameters.

\vspace{-5pt}
\paragraph{Prompt Tuning Ablation Study} To assess the effectiveness of soft
prompting, we have conducted the requested analysis and provide the results
below. Our study focused on evaluating the impact of soft prompting on the
performance of our MediSwift models during the fine-tuning phase, across both
dense and sparse configurations (see Table~\ref{tab:datasets}). The approach
involves steering the pre-trained language model by appending several additional
virtual tokens as prompts before the text. These continuous embeddings, distinct
from the main text, are randomly initialized and learned end-to-end on
downstream tasks, making them task-specific. Unlike prefix
tuning~\citep{li-liang-2021-prefix}, we strategically place the virtual tokens
not at the very beginning of the source input but specifically before the target
sequence, resulting in a final sequence structure of [source; prompt; target]
(as described in Section~\ref{subsec:prompting}). The application of soft
prompting yields a 0.7\% accuracy increase in the dense MediSwift-XL model, with
significant gains also seen in the 50\% and 75\% sparse configurations. This
highlights soft prompting's role in refining outputs through task-specific
conditioning, boosting accuracy in biomedical question answering tasks.

\begin{table}[]
  \caption{\textbf{Ablation study results evaluating the impact of soft
  prompting on MediSwift-XL models during fine-tuning.} We compared dense and
  sparse configurations (50\% and 75\% sparsity) with and without soft
  prompting. Soft prompting consistently improved accuracy across all
  configurations.}
  \begin{tabular}{lcccc}
    \hline
  \multicolumn{1}{c}{\multirow{2}{*}{\textbf{MediSwift-XL}}} &
  \multirow{2}{*}{\textbf{Size}} &
  \multirow{2}{*}{\begin{tabular}[c]{@{}c@{}}\textbf{Dense} \\
  \textbf{Fine-tune}\end{tabular}} &
  \multirow{2}{*}{\begin{tabular}[c]{@{}c@{}}\textbf{Soft} \\
  \textbf{Prompt}\end{tabular}} & \multirow{2}{*}{\textbf{Acc.}} \\
  \multicolumn{1}{c}{}                       &                       & & & \\
  \hline
  Dense                    & 1.21B                 & \checkmark & \ding{55} &
  76.1                  \\
  Dense                      & 1.21B                 & \checkmark & \checkmark &
  \textbf{76.8}                 \\ \hline
  $s=50\%$                & 605M                  & \checkmark & \ding{55} &
  75.7                  \\
  $s=50\%$              & 605M                  & \checkmark & \checkmark &
  \textbf{76.3}                  \\ \hline
  $s=75\%$                & 302M                  & \checkmark & \ding{55} &
  74.8                  \\
  $s=75\%$                & 302M                  & \checkmark & \checkmark &
  \textbf{75.4}                  \\ \hline  
  \end{tabular}
  \vspace{-5pt}
\end{table}

\vspace{-5pt}
\subsection{Document Classification on HoC} \label{subsec:doc} We examine the
Hallmarks of Cancers (HoC) corpus~\citep{Baker2016AutomaticSC}, comprising 1580
PubMed abstracts annotated for ten cancer hallmarks. We tackle a document
classification task, assigning documents to predefined single or multi-label
categories, and using MediSwift to generate label words. We follow the
established train/dev/test splits of 1108/157/315~\citep{yupubmedbert}. Similar
to~\citet{luobiogpt}, we employ a continuous embedding of length $n$ = 1 as the
prompt, and we incorporate labels into the target sequence. 

The performance is evaluated using the micro-F1 score, allowing direct
comparison with prior models and demonstrating our method's effectiveness. In
Table~\ref{tab:hoc_results}, the dense MediSwift-XL model outperformed all
similarly sized models in micro-F1 score, with its 50\% and 75\% sparse showing
very competitive results, emphasizing sparse pre-training's balance of
computational efficiency and accuracy. This further showcases the potential of
sparsity in optimizing language model performance for biomedical applications.

\begin{table}[!t]
  \begin{center}
  \caption{MediSwift's performance on the Hallmarks of Cancers (HoC) document
  classification task in both dense and sparse settings, $s \in \{50\%, 75\%\}$.
  This table compares MediSwift against other language models, demonstrating its
  efficiency-accuracy improvement. The \textit{size} column indicating
  pre-training parameters and final column reporting micro-F1 score on the test
  set.}
  \label{tab:hoc_results}
  \begin{tabular}{l@{\hspace{0pt}}cc}
    \hline
  \multicolumn{1}{c}{\textbf{Model}} & \textbf{Size} & \textbf{F1} \\\hline
  BioBERT$_{\text{\citep{lee2019}}}$                             & 110M & 81.54
  \\
  PubMedBERT$_{\text{\citep{yupubmedbert}}}$                         & 110M &
  82.32             \\
  BioLinkBERT$_{\text{base,~\citep{yasunaga-etal-2022-linkbert}}}$ & 110M &
  84.35             \\
  BioLinkBERT$_{\text{large,~\citep{yasunaga-etal-2022-linkbert}}}$ & 340M &
  84.57             \\
  GPT-2$_{\text{med,~\citep{luobiogpt}}}$                           & 345M &
  81.54             \\
  BioGPT$_{\text{med,~\citep{luobiogpt}}}$                       & 345M & 85.12
  \\
  BioGPT$_{\text{large,~\citep{luobiogpt}}}$                      & 1.54B &
  84.40             \\
  \hline \hline
  MediSwift-Med$_{\text{dense}}$                          & 302M          &
  85.15             \\
  MediSwift-Med$_{\text{s=50\%}}$                          & 151M          &
  84.48             \\
  MediSwift-Med$_{\text{s=75\%}}$                         & 0.76M         &
  83.95             \\\hline
  MediSwift-Large$_{\text{dense}}$                          & 510M          &
  85.22             \\
  MediSwift-Large$_{\text{s=50\%}}$                          & 255M          &
  84.63             \\
  MediSwift-Large$_{\text{s=75\%}}$                          & 128M          &
  84.12             \\\hline
  MediSwift-XL$_{\text{dense}}$                         & \textbf{1.21B} &
  \textbf{85.46}             \\
  MediSwift-XL$_{\text{s=50\%}}$                          & 605M          &
  84.98             \\
  MediSwift-XL$_{\text{s=75\%}}$                          & 302M          &
  84.71      \\ \hline      
  \end{tabular}
\end{center}
\vspace{-10pt}
\end{table}

\section{Related Work}

\paragraph{Sparse Training for Language Models}
Sparse weight training for language models (LM) have emerged as a promising
avenue to address the computational intensity of training large models. Recent
work has explored various sparse training
methods~\citep{thangarasa2023spdf,dao2022monarch,chen2022pixelated},
aiming to maintain or enhance model performance while significantly reducing
computational requirements. Techniques such as pruning~\citep{chen2020lottery},
sparse activations~\citep{mirzadeh2024relu}, along with the development of
specialized software~\citep{neural_magic_2021, abhaysparse2024} and
hardware~\citep{cerebrasHarnessingPower,lie_2022, dietrich2021structured} have
been pivotal. We build on these foundations, focusing on optimizing
sparse weight training strategies specifically for domain-specific LMs, pushing
the boundaries of efficiency.

\vspace{-5pt}
\paragraph{Biomedical Language Models} 
The evolution of language models for medical applications has progressed from
adapting encoder-only architectures like BERT~\citep{devlin2018bert}, using
biomedical corpora~\citep{lee2019, yupubmedbert}, to incorporating strategies
like document links~\citep{yasunaga-etal-2022-linkbert} and knowledge
graphs~\citep{yasunaga2022dragon}. The shift towards autoregressive generative
models, such as GPT~\citep{brown2020language} and
Llama~\citep{touvron2023llama}, for pretraining on medical texts has led to
significant advancements~\citep{wu2023pmcllama, luobiogpt, elliotbiomedlm}.
Recent scaling efforts include GatorTronGPT with 20B
parameters~\citep{Yang2022gatortron}, as well as
Clinical-Camel~\citep{toma2023clinical}, MEDITRON~\citep{chen2023meditron70b}
and Med-42~\citep{med42}, based on Llama-2-70B~\citep{touvron2023llama2},
focusing on mixed clinical and general English texts. Our work diverges from
works that scaled up medical LMs by introducing weight sparsity into the
pre-training of biomedical LMs. This reduces the computational costs typically
associated with large-scale models, thereby improving the balance between
efficiency and accuracy in the medical domain.

\vspace{-5pt}
\paragraph{Prompting for Biomedical Language Models}
Recent research has shifted towards prompt engineering to enhance language
models' performance on biomedical tasks, such as BioGPT's~\citep{luobiogpt} use
of soft prompt-tuning and Medprompt's~\citep{nori2023generalist} innovative
prompting techniques for generalist foundation models. \citet{lievin2023large}
and \citet{yagnik2024medlm} analyzed the effectiveness of prompting in the
medical domain and showed that it can improve metric scores. However, combining
task-specific fine-tuning with prompting strategies, as seen in Med-PaLM
2~\citep{singhal2022large,singhal2023expertlevel}, yields competitive results on
challenging biomedical tasks. Our work extends this by integrating task-specific
fine-tuning and soft prompting to address model accuracy loss during sparse
pre-training, effectively achieving efficiency gains with minimal accuracy
degradation on biomedical tasks.

\vspace{-5pt}
\section{Conclusion}

In conclusion, MediSwift innovates in biomedical language models by combining
sparse pre-training with dense fine-tuning and soft prompting, balancing
computational efficiency with accuracy. Available in Med, Large, and XL sizes,
with 50\% and 75\% sparsity, MediSwift addresses the cost of training models and
sets new standards for biomedical tasks like PubMedQA. MediSwift-XL, in
particular, showcases superior efficiency-accuracy trade-offs, outperforming
models up to 7B parameters. This work exemplifies the potential of sparse
pre-training as a cost-effective method for developing specialized,
high-performance models, establishing MediSwift as a benchmark in biomedical
NLP.

\section*{Limitations}
Our work on MediSwift represents a significant leap forward in developing
efficient domain-specific LLMs, particularly in biomedicine, by utilizing sparse
pre-training to strike a fine balance between computational efficiency and
accuracy. While we have initially focused on static sparse pre-training, the
emerging field of dynamic sparse training (DST) holds great promise for further
improvements~\citep{evci2020rigging,mocanu2018,liu2021sparse}. 

DST offers an exciting avenue for optimizing sparsity patterns dynamically,
potentially elevating model quality and training efficiency to new heights.
Although the implementation of DST requires advanced software and hardware
support for unstructured sparse computations—capabilities that were beyond our
current scope—this innovative approach represents an interestingly opportunity
for future research. As support for unstructured sparse training evolves with ML
software-hardware co-design, we anticipate these advancements will enable us to
harness DST, paving the way for even more high-quality and efficient
domain-specific LLMs.

\section*{Ethics Statement}
While MediSwift represents a significant advancement in encoding medical
knowledge from sources of high-quality evidence, it is important to note that it
has not been fully adapted to deliver this knowledge in a manner that is
appropriate, safe, or within the actionable constraints required by medical
professionals. Therefore, we strongly recommend against deploying MediSwift
directly in clinical or medical applications without thorough alignment with
specific use cases. 

Moreover, additional testing is crucial, including the conduct of randomized
controlled trials in real-world practice settings, to ensure the model's
recommendations are reliable and beneficial in practical healthcare
environments. This cautionary approach emphasizes the importance of bridging the
gap between technological capabilities and the nuanced requirements of medical
practice to ensure patient safety and efficacy of care.

\bibliography{custom}
\clearpage
\appendix

\section{Experimental Setup and Hyperparameter Details}
\label{app:hyperparameters}

\subsection{Pre-training on Biomedical Data} \label{appsub:pretrain} To train
all MediSwift models, we use the AdamW optimizer~\citep{loshchilov2017decoupled}
with a peak learning rate set at 2$\times$10$^{\text{-4}}$, $\beta_1 = $0.9,
$\beta_2 = $0.95 and $\epsilon = $10$^{\text{-8}}$. A linear warmup period,
amounting to 10\% of the total training steps, is employed before transitioning
to a cosine decay schedule, with the learning rate decreasing to a minimum of
10\% of the peak value (i.e., 2$\times$10$^{\text{-5}}$). In
Table~\ref{app:archs}, we provide details on the size and architecture
configurations of the MediSwift models we pre-trained. Here, $n_{\text{params}}$
is the total number of trainable parameters, $n_{\text{layers}}$ is the number
of decoder layers, and $d_{\text{model}}$ is the base size of the model. The
feedforward bottleneck is four times the base size, i.e., $d_{\text{ff}}$ = 4 ×
$d_{\text{model}}$. Finally, $n_{\text{heads}}$  are the number of attention
heads and $d_{\text{head}}$ is the dimension of each attention head. The context
window size is set to 1024.

\begin{table}[!ht]
  \setlength{\tabcolsep}{7pt}
  \centering
  \caption{ Sizes, architectures, and pre-training hyperparameters (batch size,
  learning rate, etc.) of the MediSwift models at three sizes (i.e., Med, Large
  and XL), which are trained for a total of 104.86B tokens.}
  \label{app:archs}
  \begin{tabular}{lccc}
    \hline
    & \multicolumn{3}{c}{\textbf{MediSwift Models}} \\ \cline{2-4} &
    \multicolumn{1}{c}{Med} & \multicolumn{1}{c}{Large} & \multicolumn{1}{c}{XL}
    \\ \hline
  $n_{\text{params}}$                                              & 302M & 510M
  & 1.21B                 \\
  $n_{\text{layers}}$                                              & 24 & 18 &
  24          \\
  $d_{\text{model}}$                                               & 1024 & 1536
  & 2048              \\
  $n_{\text{heads}}$                                               & 16 & 12 &
  16            \\
  $d_{\text{head}}$                                                & 64 & 128 &
  128                     \\ \hline
  \hline
  Batch Size                                             &
  \multicolumn{3}{c}{512}                                 \\
  MSL                                            & \multicolumn{3}{c}{1024} \\
  Optimizer & \multicolumn{3}{c}{AdamW}   \\
  Warmup Schedule & \multicolumn{3}{c}{Linear} \\ 
  Decay Schedule & \multicolumn{3}{c}{Cosine} \\ 
  LR                                                     &
  \multicolumn{3}{c}{2$\times$10$^{\text{-4}}$}                            \\
  Weight Decay & \multicolumn{3}{c}{0.1}    \\

  Total Steps & \multicolumn{3}{c}{200,000}   \\
  Warmup Tokens & \multicolumn{3}{c}{10.486$\times$10$^{\text{9}}$} \\
  Training Tokens & \multicolumn{3}{c}{104.86$\times$10$^{\text{9}}$} \\ \hline
  \end{tabular}
\end{table}

Following the training FLOPs calculation described in~\citet{hoffmann2022an}, we
compute the total pre-training FLOPs for the dense and sparse variants of
MediSwift-Med, Large and XL, and report them in
Table~\ref{tab:mediswift_flops_ext}, along with their relative FLOPs reduction
over the dense baseline. Similar to Appendix F of~\citet{hoffmann2022an}, we
also include the training FLOPs contributed by the embedding matrices.
Additionally, in large models, the contribution of embedding matrices to the
overall FLOPs and parameters is minimal. 

\begin{table}[t]
  \centering
  \caption{\textbf{Final Pre-training Losses and Computational Efficiency of
  MediSwift Models.} We summarize the results for the MediSwift-Med and
  MediSwift-XL models, trained using the biomedical pre-training corpus detailed
  in Section~\ref{sec:pretraincorpa}. We report the final pre-training losses
  for dense and sparse variants at 50\% and 75\% sparsity levels. The total
  FLOPs and FLOP savings relative to the dense baseline are indicated in
  parentheses, highlighting the models' computational efficiency.}
  \label{tab:mediswift_flops_ext}
  \begin{tabular}{lccc}
    \hline
    \textbf{MediSwift} & \textbf{Size} &
    \textbf{\begin{tabular}[c]{@{}c@{}}Pre-train \\ Loss\end{tabular}} &
    \textbf{\begin{tabular}[c]{@{}c@{}}Train \\ FLOPs (x 10$^{20}$)
    \end{tabular}} \\ \hline
    Med$_{\text{dense}}$ & 302M & 2.234 & 2.677 (1.00x) \\
    Med$_{s=50\%}$ & 151M & 2.265 & 1.727 (0.64x) \\
    Med$_{s=75\%}$ & 0.76M & 2.375 & 1.252 (0.46x) \\ \hline \hline
    Large$_{\text{dense}}$ & 510M & 2.047 & 4.248 (1.00x) \\
    Large$_{s=50\%}$ & 255M & 2.172 & 2.645 (0.62x) \\
    Large$_{s=75\%}$ & 128M & 2.281 & 1.840 (0.43x)                \\ \hline
    \hline
    \hline
    XL$_{\text{dense}}$ & 1.21B & 1.979 & 9.148 (1.00x) \\
    XL$_{s=50\%}$ & 605M & 2.012 & 5.348 (0.58x) \\
    XL$_{s=75\%}$ & 302M & 2.141 & 3.448 (0.38x)                \\ \hline                                  
    \end{tabular}
\end{table}

\subsection{PubMedQA Fine-tuning}\label{appsec:pubmedqa}

As mentioned in Section~\ref{subsec:pubmedqa}, the PubMedQA dataset includes
three subsets: PQA-A, PQA-U, and PQA-L. We train all of our MediSwift models on
the original train/val/test splits for each of these datasets in a multi-stage
manner~\citep{jin2019pubmedqa}, both dense and sparse, using
AdamW~\citep{loshchilov2017decoupled} and a linear learning rate warmup (i.e.,
10\% of to the total training steps) followed by a cosine decay schedule for a
maximum 5 epochs, and perform early-stopping when the models began to overfit.
We perform a grid search to discover an appropriate learning rate that led to
the best downstream classification accuracy on each of the three datasets for a
given compute budget. More specifically, on the dense baseline and sparse
variants, we select the best batch size among $\{$8, 16, 32, 64$\}$ and select
the best learning rate among $\{$2e-4, 1e-4, 5e-5, 2.5e-5$\}$ on the validation
set. After training on the final stage (i.e., PQA-L), we evaluate the model on
the PQA-L test set using the official evaluation
scripts\footnote{\url{https://github.com/pubmedqa/pubmedqa}}. All results were
averaged across 3 random seeds.

\subsection{HoC Fine-tuning}\label{appsec:hoc}

In Section~\ref{subsec:doc}, we described the Hallmarks of Cancers (HoC) dataset
which comprises of 1580 PubMed abstracts with a 1108/157/315 split for train,
val and test sets following~\citet{yupubmedbert}. We train all of our MediSwift
models on the original train/val/test splits for both dense and sparse, using
AdamW~\citep{loshchilov2017decoupled} and a linear learning rate warmup (i.e.,
10\% of to the total training steps) followed by a cosine decay schedule for a
total of 100 epochs, and perform early-stopping when the models began to
overfit. We perform a grid search to discover an appropriate learning rate that
led to the best micro-F1 score for a given compute budget. More specifically, on
the dense baseline and sparse variants, we select the best batch size among
$\{$16, 32, 64$\}$ and select the best learning rate among $\{$8e-5, 4e-5, 2e-5,
1e-5$\}$ on the validation set. All results were averaged across 3 random seeds.

\begin{figure}[]
      \centering
      \includegraphics[width=\linewidth]{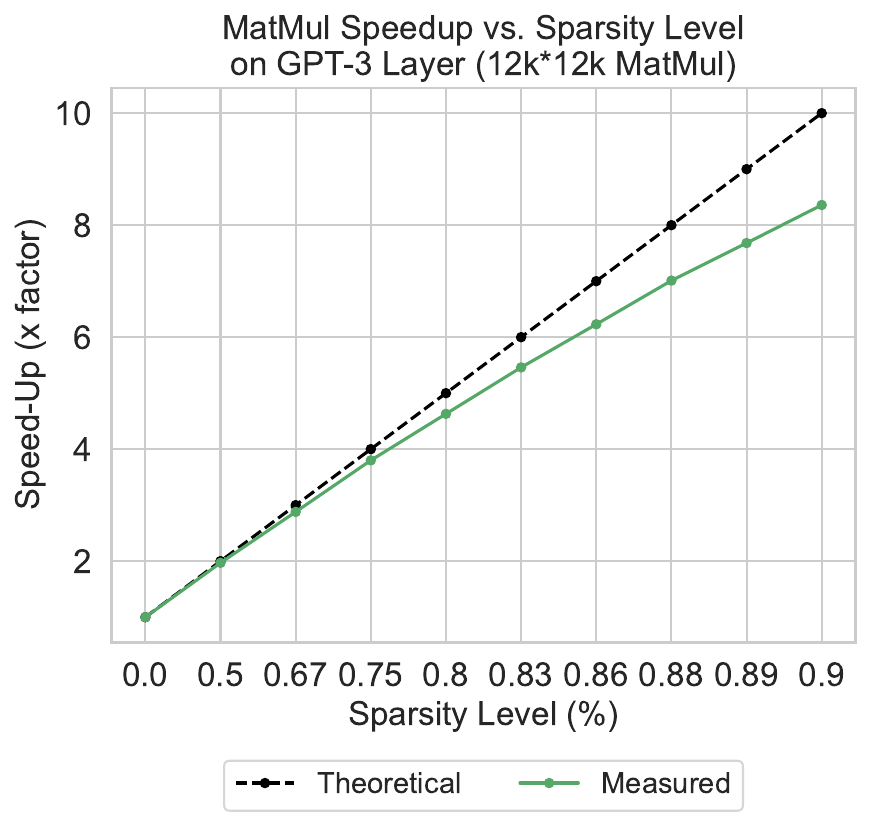}
      \caption{Comparison of measured speedup versus theoretical speedup for
      GPT-3 layer 12k $\times$ 12k matrix multiplication (MatMul) on the
      Cerebras CS-2 system at various sparsity levels. This graph illustrates
      the efficiency gains achieved through sparse computation, highlighting the
      real-world performance relative to theoretical predictions.}
      \label{fig:gpt3_speedup}
\end{figure}

\section{Unstructured Sparsity on Specialized Hardware
Accelerators}\label{appsec:cs2} The Cerebras CS-2 system, designed specifically
for accelerating deep learning computations, can handle unstructured sparsity
efficiently due to its unique
architecture~\citep{cerebrasHarnessingPower,lie_2021}. The CS-2's wafer-scale
engine, with its vast array of computational cores and on-wafer memory,
efficiently manages unstructured sparsity's irregular memory access, surpassing
traditional architectures that often face memory bandwidth constraints.
Moreover, the CS-2 has a significant amount of on-chip memory, reducing the need
to access external memory. This is crucial for unstructured sparsity, as the
irregular access patterns can lead to high latency if data needs to be fetched
from off-chip. By keeping more data on-chip, the CS-2 minimizes these latencies.
In addition, the system supports fine-grained parallelism, allowing it to
execute many small, sparse operations concurrently across its thousands of
cores. This is particularly advantageous for unstructured sparsity, as the
workload can be distributed across many cores to maintain high utilization. In
Figure~\ref{fig:gpt3_speedup}, we highlight the potential realized gains with
unstructured weight sparsity on the Cerebras CS-2.

\section{Author Contributions}

We provide a summary of each author's contributions:
\begin{itemize}
  \item \textbf{Vithursan Thangarasa} led the efforts on developing the method and pre-training the various dense and sparse MediSwift models. He also implemented several parts of the code for fine-tuning and proposed innovative approaches such as dense fine-tuning and soft prompting. Additionally, he wrote the manuscript, ensuring a comprehensive and cohesive presentation of the research.
  
  \item \textbf{Mahmoud Salem} assisted in running experiments for both single-phase and multi-phase fine-tuning on PubMedQA. His work was crucial in validating the effectiveness of the proposed methods across different fine-tuning strategies.
  
  \item \textbf{Shreyas Saxena} collaborated closely with Vithursan to help design the main method. His insights and feedback were vital in refining the overall approach and ensuring the robustness of the proposed techniques.
  
  \item \textbf{Kevin Leong} contributed to the pre-training of both dense and sparse MediSwift models on the Cerebras CS-2. His expertise in utilizing the Cerebras CS-2 hardware was essential for the efficient and effective training of the models.
  
  \item \textbf{Joel Hestness} was involved in discussions with Vithursan and provided valuable feedback to improve the manuscript. His contributions helped enhance the clarity of the manuscript.
  
  \item \textbf{Sean Lie} frequently met with Vithursan for technical discussions and provided support for bringing up the Cerebras CS-2. His involvement was critical in performing the pre-training of MediSwift models from scratch and ensuring the smooth operation of the hardware.
  \end{itemize}  
\end{document}